\documentclass[conference]{IEEEtran}
\IEEEoverridecommandlockouts
\usepackage{cite}
\usepackage{amsmath,amssymb,amsfonts}
\usepackage{graphicx}
\usepackage{textcomp}
\usepackage{xcolor}
\usepackage{hyperref}
\usepackage{multicol}
\usepackage{algorithm,algpseudocode}
\usepackage{subcaption}
\usepackage[left=1.62cm,right=1.62cm,top=1.9cm,bottom=2.7cm]{geometry}

\usepackage{times}
\usepackage{epsfig}
\usepackage{nicefrac}
\usepackage{mathtools}
\usepackage{adjustbox}

\usepackage{caption}
\usepackage{tikz}
\usepackage{pgfplots}
\usetikzlibrary{spy,calc}
\usepackage{array}

\newcolumntype{x}[1]{>{\centering\arraybackslash\hspace{0pt}}p{#1}}

\def\BibTeX{{\rm B\kern-.05em{\sc i\kern-.025em b}\kern-.08em
    T\kern-.1667em\lower.7ex\hbox{E}\kern-.125emX}}
\begin{document}

\title{A Scalable Decentralized Reinforcement Learning Framework for UAV Target Localization Using Recurrent PPO \\
}

\author{\IEEEauthorblockN{Leon Fernando\IEEEauthorrefmark{1},
Billy Lau Pik Lik\IEEEauthorrefmark{2},
Chau Yuen\IEEEauthorrefmark{3},
U-Xuan Tan\IEEEauthorrefmark{4}}
\IEEEauthorblockA{\IEEEauthorrefmark{1}Department of Electronic and Telecommunication Engineering, University of Moratuwa, Sri Lanka, fernandoknal.20@uom.lk}
\IEEEauthorblockA{\IEEEauthorrefmark{2}Engineering Product Development, Singapore University of Technology and Design, Singapore, billy\_lau@sutd.edu.sg}
\IEEEauthorblockA{\IEEEauthorrefmark{3}School of Electrical and Electronic Engineering, Nanyang Technological University, Singapore, chau.yuen@ntu.edu.sg}
\IEEEauthorblockA{\IEEEauthorrefmark{4}Engineering Product Development, Singapore University of Technology and Design, Singapore, uxuan\_tan@sutd.edu.sg}
}

\maketitle

\begin{abstract}

The rapid advancements in unmanned aerial vehicles (UAVs) have unlocked numerous applications, including environmental monitoring, disaster response, and agricultural surveying. Enhancing the collective behavior of multiple decentralized UAVs can significantly improve these applications through more efficient and coordinated operations. In this study, we explore a Recurrent PPO model for target localization in perceptually degraded environments like places without GNSS/GPS signals. We first developed a single-drone approach for target identification, followed by a decentralized two-drone model. Our approach can utilize two types of sensors on the UAVs, a detection sensor and a target signal sensor. The single-drone model achieved an accuracy of 93\%, while the two-drone model achieved an accuracy of 86\%, with the latter requiring fewer average steps to locate the target. This demonstrates the potential of our method in UAV swarms, offering efficient and effective localization of radiant targets in complex environmental conditions.

\end{abstract}
\vspace{0.2cm}
\begin{IEEEkeywords}
Target localization, Unmanned aerial vehicle (UAV), Artificial Intelligence (AI), Reinforcement learning (RL), Multi-Agent Recurrent PPO
\end{IEEEkeywords}

\section{Introduction}
\label{sec:Introduction}
Unmanned aerial vehicles (UAVs) have gained increasing interest due to their rapid evolution and expanding applications in various fields such as environmental monitoring \cite{s17102234}, disaster response, agricultural surveying \cite{Alsalam2017AutonomousUW} and search and rescue (SAR) operations\cite{Atif2021UAVAssistedWL}. UAVs offer several advantages, including cost efficiency, flexibility, mobility, and lightweight design, making them invaluable for tasks like target identification, tracking, data collection, delivery, and communication services\cite{7995044},\cite{DBLP:journals/corr/abs-1802-07187}, \cite{article8}.  Additionally, there is a growing focus on developing UAVs equipped with sensors for Smart Environmental Monitoring (SEM) applications as discussed in \cite{SHURRAB2023100867}.  

Target localization, particularly signal-based target identification is a critical aspect of SEM. It involves determining the position of an unknown radiant target within an area of interest (AOI) using sensor readings, without prior knowledge of the target's characteristics, such as location. This target can be any physical phenomenon, including radiation, fires, or pollution \cite{article11},\cite{article13}. Furthermore, in SAR operations, quickly locating individuals based on signals from distress beacons, mobile phones, or GPS devices can be lifesaving, enabling the rapid localization of lost hikers, disaster survivors, or stranded individuals.

UAVs have evolved as crucial tools for target localization. Integrating AI, including reinforcement learning (RL) \cite{sutton2018reinforcement}, has greatly enhanced UAVs' autonomy and intelligence in target detection. This advancement allows UAVs to learn optimal decision-making strategies from their experiences in uncertain and dynamic environments. Despite these innovations, deploying UAVs for such tasks presents challenges and limitations. In complex environments like mining, and SAR operations, challenges such as GNSS/GPS signal absence and limited visibility severely restrict UAV capabilities. These constraints are exacerbated in indoor settings where signal denial and reduced visibility limit information availability. Boiteau et al. \cite{rs16030471} propose a solution based on Partially Observable Markov Decision Processes (POMDPs) using thermal cameras for search tasks under varying visibility conditions. Operating UAVs at low altitudes in cluttered environments poses challenges for GPS-based navigation, making alternatives with reduced hardware and power requirements necessary \cite{Aasish2015NavigationOU}.

To address these limitations, we propose a data-driven autonomous UAV target localization system designed for versatile environments, with particular efficacy in challenging indoor conditions as discussed above. Our system uses a Recurrent PPO model by integrating Long Short-Term Memory (LSTM) networks with Proximal Policy Optimization (PPO) algorithms. The strength of our approach lies in its ability to enable UAVs to localize objects using limited information from the environment and using minimal, less complex accessories, such as a single signal tracking sensor and a detection sensor (e.g. depth sensor or camera). 
\\The key contributions of this paper include: 
\begin{enumerate}
    \item A generalized approach to target localization for a single-drone in perceptually degraded environments.
    \item A generalized decentralized and scalable approach to target localization for multiple drones in perceptually degraded environments.
\end{enumerate}

In the remainder of the paper, we detail the methodology of the proposed system in Section \ref{sec:Methodology}. Section \ref{sec:RD} presents the results of our work along with a discussion of the findings. Finally, we conclude with suggestions for future improvements in Section \ref{sec:CF}.

\section{Methodology}
\label{sec:Methodology}
In this section, we introduce our system's overview, detailing the observation space, action space, reward system, model selection, and training methodology. Our system aims in enabling drones to efficiently navigate an environment to locate a target, emitting a signal at a specific altitude. 

\subsection{Observation Space}
The observation space is the input for the model that the drone perceives from its surroundings. The simulation environments enable the drone to create a searched grid map based on its movements. (Multi-drone environment creates a single shareable grid map among the swarm members.) The grid map consists of the states as in Table \ref{tab:States}

\begin{table}[htbp]
\caption{States for the grid map for both simulation environments}
\begin{center}
\begin{tabular}{|c|p{6cm}|}
\hline
\textbf{State} & \textbf{Description} \\
\hline
0 & Unknown cell (have not been explored) \\
\hline
1 & Obstacle \\
\hline
2 & Once travelled cell \\
\hline
3 & Twice travelled cell \\
\hline
4 & Thrice or more travelled cell\\
\hline
\end{tabular}
\label{tab:States}
\end{center}
\end{table}

The observation states of the single drone simulation environment consists of 17 inputs. When scaling into multiple drones, additional 4 signals are included in the observation space to check for nearby drones, totaling upto 21 inputs.
Those are summarized in Table \ref{tab:observations}.

\begin{table}[h!]
    \caption{Observation Inputs for Single Drone and Multi-Drone Simulation Environment}
    \centering
    \begin{tabular}{| >{\bfseries}m{1.5cm} | m{4.5cm} | m{1cm} |}
        \hline
        \textbf{Category} & \textbf{Details} & \textbf{Inputs} \\
        \hline
        \multicolumn{3}{|c|}{\textbf{For both single and multi drone simulation environments}} \\
        \hline
        Surrounding Directions & Presence of obstacles in the 8 surrounding directions & 8 \\
        \hline
        Map States & States of the map in the 8 surrounding directions & 8 \\
        \hline
        Signal Strength & Strength of the signal emitted by the target & 1 \\
        \hline
        \multicolumn{3}{|c|}{\textbf{Additional Inputs for Multi-drone Simulation environment.}} \\
        \hline
        Neighbouring drones & Presence of drones in vicinity of
        the 4 quadrants. & 4 \\
        \hline
    \end{tabular}
    \label{tab:observations}
\end{table}

\begin{table}[h!]
    \caption{Signal Strength Sections}
    \centering
    \begin{tabular}{| >{\bfseries}m{1.5cm} | m{5.5cm} |}
        \hline
        \textbf{Section} & \textbf{Details} \\
        \hline
        Section 1 & Highest signal strength, reaching this is considered reaching the target, terminating the task \\
        \hline
        Section 2 & Moderate signal strength\\
        \hline
        Section 3 & Weak signal, beyond which the signal is not available \\
        \hline
    \end{tabular}
   
    \label{tab:signal_strength}
\end{table}
The signal emitted by the target is divided into three sections as in Table \ref{tab:signal_strength}. This signal strength classification serves as an indicator for the model, specifying the proximity of the target within the weak signal radius. UAVs are unable to detect any signal from the target beyond this weak signal radius. The signal strength for multi drone environment is determined in the same way as in a single-drone environment.
\vspace{0.2cm}

\begin{figure}[h!]
        \centering
        \includegraphics[width=0.8\linewidth]{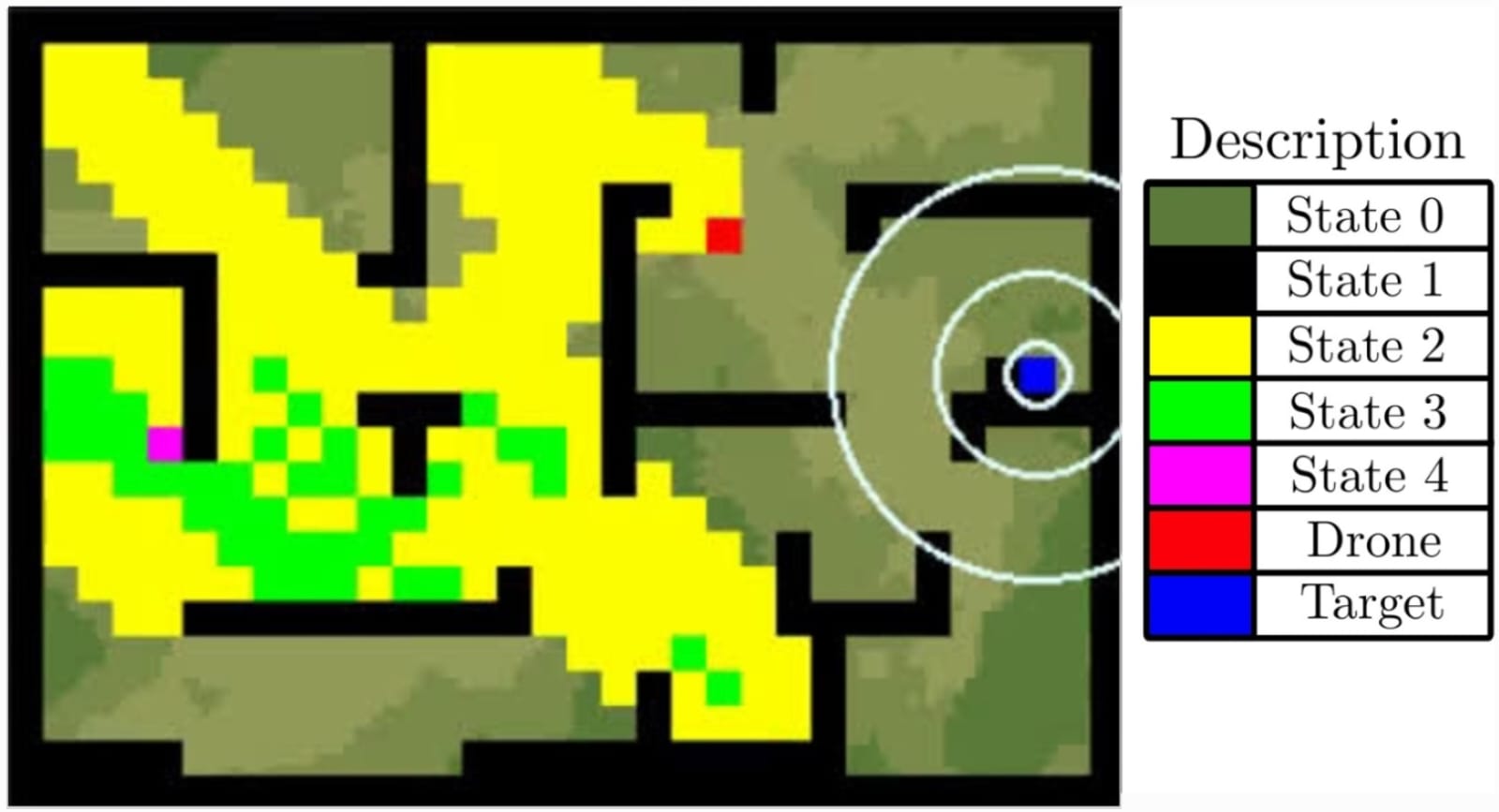}
    \caption{Grid map representation of the Single drone simulation environment. The unexplored cells are in state 0, the explored once cells in state 2 and yellow, explored twice cells in state 3 and green, explored thrice or more cells in state 4 and pink. The obstacles are represented in black, in state 1.}
    \label{fig:grid_map}
\end{figure}
\subsection{Action Space}
The actions that the UAV can take to move in the simulation environment are called the action space. For both simulation environments, the action space of the drone includes movement in eight directions. For simplification, we restrict actions to pure translation in these eight directions.

\subsection{Reward System}
In RL, the reward system provides feedback to the agent about its actions. The reward system incentivizes exploration and efficient navigation toward the target while penalizing undesirable actions. For the multi-drone scenario, additional considerations were incorporated into the reward system to manage interactions between drones. The reward structures are summarized in Table \ref{tab:reward_model}.

\begin{table}[h!]
\caption{Reward Model for Single and Multi-Drone Scenarios}
\centering
\begin{tabular}{|p{4cm}|p{1.4cm}|p{1.4cm}|}
\hline
\textbf{Description}&\multicolumn{2}{|c|}{\textbf{Score}} \\

\cline{2-3} 
\textbf{} & \textbf{\textit{Single }}& \textbf{\textit{Multi}} \\

\hline
Moving to Unknown Cell (State 0) & 2 & 2 \\

Moving to Obstacle Cell (State 1) & -50 & -50 \\
Moving to a cell that has been travelled once (State 2) & 0 & 0 \\
Moving to a cell that has been travelled twice (State 3) & -1 & -1 \\
Moving to a cell that has been travelled thrice/more (State 4) & -4 & -4 \\
Signal Strength based on cost function & 0.0 – 250.0 & 0.0 – 250.0 \\
Neighboring drone in the vicinity & N/A & -2 \\
Colliding with a neighboring drone & N/A & -50 \\
Target Reach Reward & 1000 & 1000 \\
\hline
\end{tabular}
\label{tab:reward_model}
\end{table}

\subsection{Model Selection}
\begin{figure}[h!]
    \centering
    \includegraphics[width=0.8\linewidth]{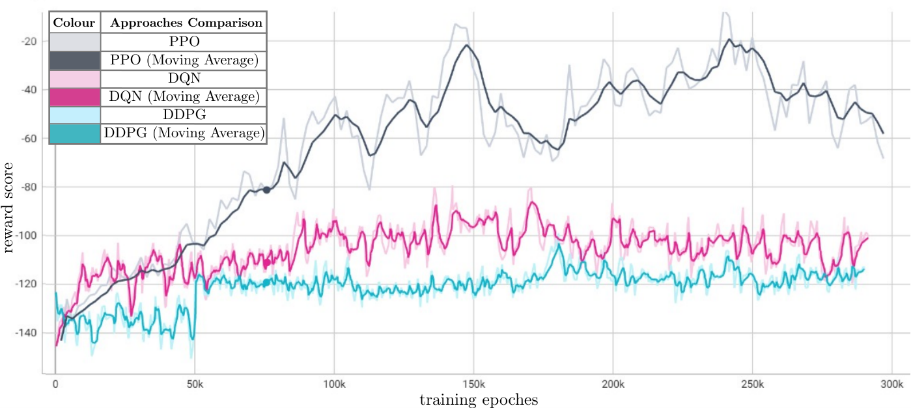}
    \caption{Model performance comparison}
    \label{fig:model_performance}
\end{figure}

Given the limited perception of drones, identifying an effective reinforcement learning model was crucial. Few well-known models such as DQN \cite{mnih2013playing}, DDPG \cite{lillicrap2019continuous}, and PPO \cite{schulman2017proximal} were chosen to compare their effectiveness with the limited information in target localization. It was found that the PPO model yielded the highest mean reward, as shown in Fig. \ref{fig:model_performance}. Therefore, the PPO model was selected as the base model for the simulation environments.

\subsection{Proposed Model Architecture}
To enhance the selected PPO model, an LSTM layer with 256 hidden units was incorporated, utilizing a Recurrent PPO model as described in \cite{pleines2022generalization}. This addition aims to improve the model's handling of partial Markov Decision Processes (MDPs). Any RL algorithm's performance can be significantly be degraded due to the agent's inability to fully capture the temporal dependencies and the underlying state dynamics. This often results in the drone getting stuck in local minima, as it fails to learn an effective way that considers the history of past observations and actions. Therefore, adding the LSTM layer enhances its overall performance in the simulation environment. The loss function of this model is calculated as in \ref{eq: PPO loss}.
\begin{equation}
\mathcal{L}^{\text{CLIP+VF+S}}(\theta) = \hat{\mathbb{E}}_t \mathcal{L}^{\text{CLIP}}_t(\theta) - c_1 \mathcal{L}^{\text{VF}}_t(\theta) + c_2 S[\pi_\theta](s_t)
\label{eq: PPO loss}
\end{equation}

\(\mathcal{L}^{\text{CLIP}}(\theta)\) denotes the clipped surrogate objective. \(c_1\) and \(c_2\) are coefficients. \(\mathbb{E}_t\) denotes the empirical expectation over timesteps.  \(\mathcal{L}^{\text{VF}}_t(\theta)\) is the Value function, which is a squared loss error. \(\mathcal{S}[\pi_\theta](s_t)\) is the entropy bonus of the policy.
    \begin{equation}
    \mathcal{L}^{\text{CLIP}}(\theta) = \hat{\mathbb{E}}_t \left[ \min \left( r_t(\theta) \hat{A}_t, \text{clip}(r_t(\theta), 1 - \epsilon, 1 + \epsilon) \hat{A}_t \right) \right]
    \label{eq: Lclip}
    \end{equation}

     \begin{equation}
    r_t(\theta) = \frac{\pi_\theta(a_t \mid o_t, h_t)}{\pi_{\theta_{\text{old}}}(a_t \mid o_t, h_t)}
    \label{eq: PR}
\end{equation}

     \(\mathcal{L}^{\text{CLIP}}(\theta)\) calculated as in \ref{eq: Lclip}, where \ref{eq: PR} is the probability ratio between the new and old policies. As in \cite{pleines2022generalization}, when using a recurrent layer like LSTM, the choice of action \( a_t \) by the policy hinges on the current observation \( o_t \) and the hidden state \( h_t \) of this recurrent layer. \(\hat{A}_t\) is determined through Generalized Advantage Estimation (GAE). \(\mathcal{L}^{\text{VF}}_t(\theta) \) and  \(\mathcal{S}[\pi_\theta](s_t)\) are calculated as same in PPO model.

\begin{figure}[h!]
    \centering
    \includegraphics[width=0.9\linewidth]{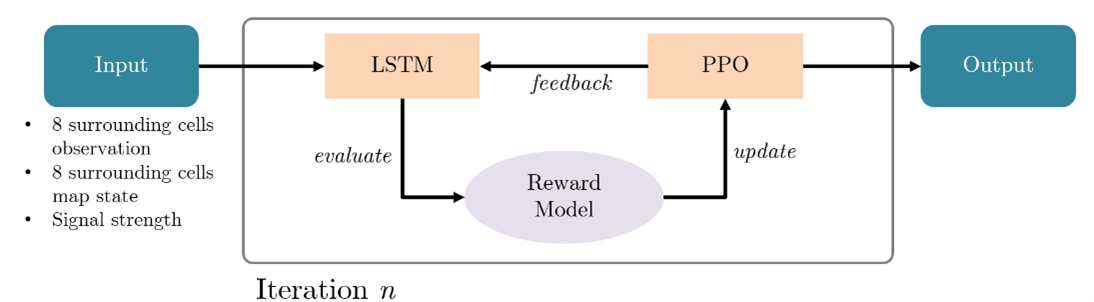}
    \caption{Proposed model architecture}
    \label{fig:model_architecture}
\end{figure}

The proposed model architecture is shown in Fig. \ref{fig:model_architecture}.\\

\subsection{Training Process}
For both single-drone and multi-drone scenarios, the training involved creating diverse datasets through eight different simulation environments. By varying the positions of the drones and the targets, 80 distinct maps with different sizes were generated. In the multi-drone training, two drones were placed instead of one. The simulation environments were altered every 50 resets, and training continued for several million epochs.

\begin{figure}[h!]
    \centering
    \begin{subfigure}{0.15\textwidth}
        \centering
        \includegraphics[width=1\linewidth]{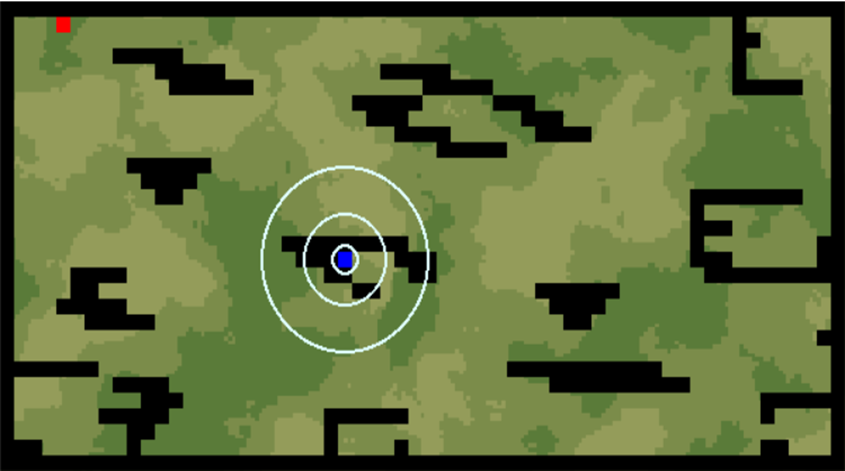}
        \caption{}
        \label{subfig:suba}
    \end{subfigure}
    \hspace{0.05\textwidth} 
    \begin{subfigure}{0.15\textwidth}
        \centering
        \includegraphics[width=1\linewidth]{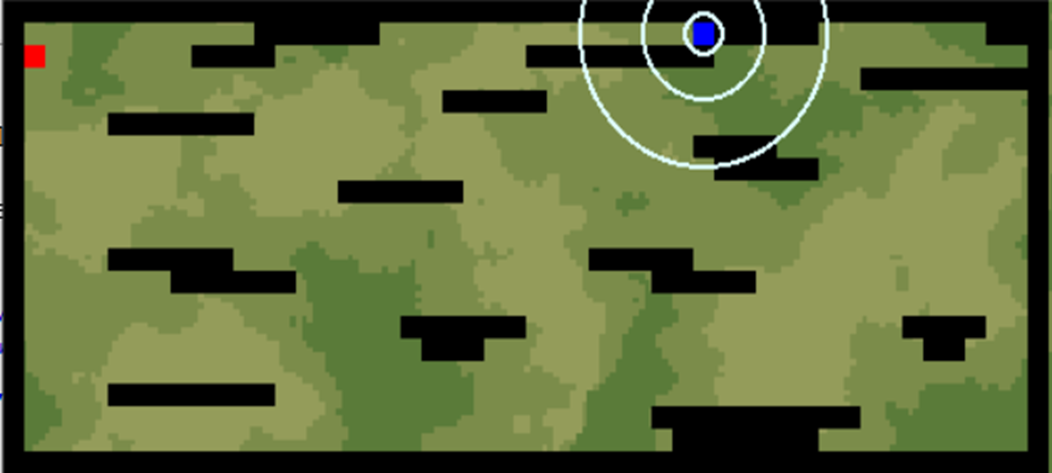}
        \caption{}
        \label{subfig:subb}
    \end{subfigure}
    \caption{Training maps for simulation environments.}
    \label{fig:Training maps}
\end{figure}


\section{Results and Discussion}
\label{sec:RD}
\begin{figure}[h!]
    \centering
    \includegraphics[width=0.8\linewidth]{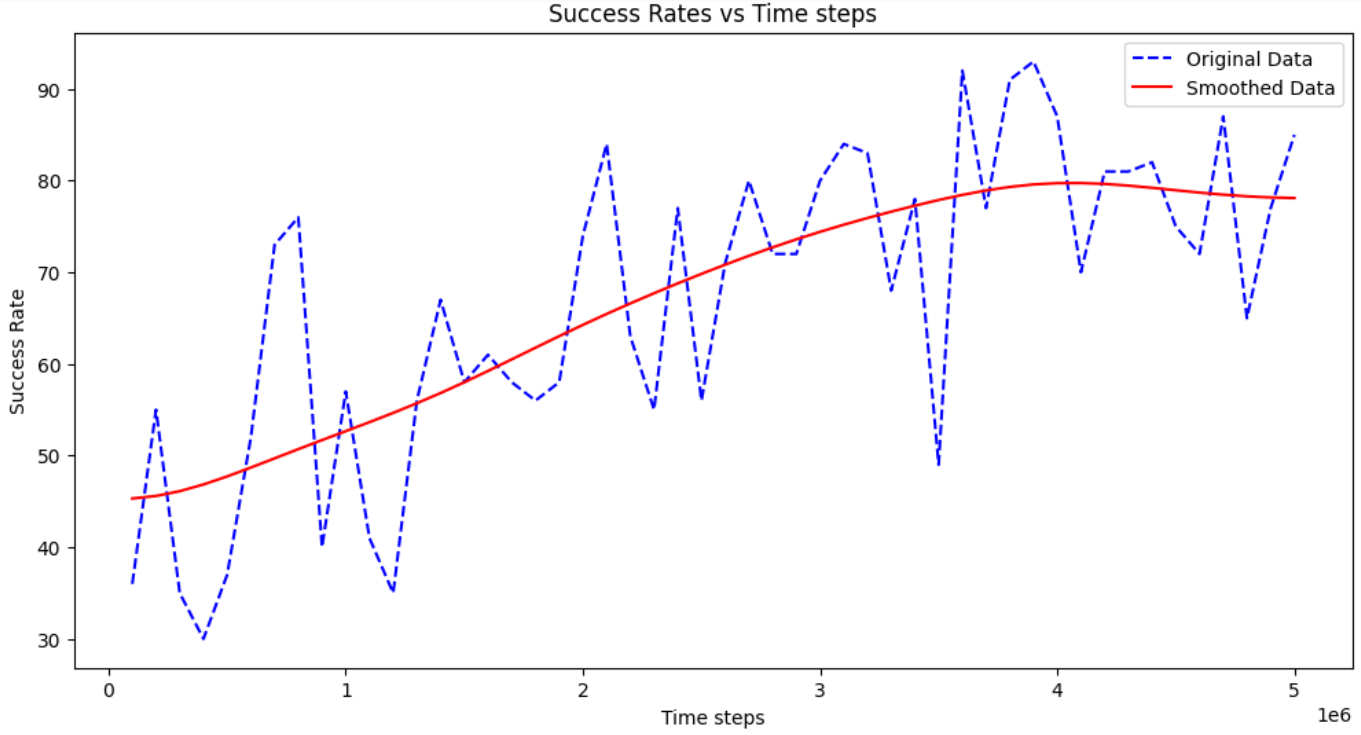}
    \caption{Evaluation graph - single drone simulation environment.}
    \label{fig:single_drone_eval}
\end{figure}

In the testing process, 10 different simulation maps were created, each depicting the indoor layout of a building. By varying the positions of the drones and the target, we generated 100 unique simulation environments. In each scenario, reaching the target was considered a success.
The success rate, \( r \), can be calculated as in \ref{eq:success}.
\begin{equation}
    r = \frac{h}{100}
    \label{eq:success}
\end{equation}
where \( h \) denotes the number of successful target localizations.
Subsequently, we want to calculate the average success rate \(\bar{r}\) as shown in \ref{eq: avg}.
\begin{equation}
    \bar{r} = \frac{T_{q}}{s}
    \label{eq: avg}
\end{equation}
\( T_{q} \) represents the total steps taken for all successful target localizations, $s$ represents the number of successful targets localizations.
The best results and number of successful target localizations versus the number of time steps (epochs) plotted are summarized in the Tables \ref{tab:single_drone_performance} and \ref{tab:multi_drone_performance}.\\
\begin{table}[h!]
\centering
\caption{Model performance for single drone}
\begin{tabular}{|l|c|c|}
\hline
\textbf{Training Iteration} & \textbf{Success Rate} & \textbf{Average Steps} \\
\hline
3.8 mil & 93\% & 183.82 \\
4.3 mil & 93\% & 185.76 \\
\hline
\end{tabular}
\label{tab:single_drone_performance}
\end{table}
\figurename~\ref{fig:single_drone_eval} shows the evaluation graph for the single drone simulation environment. The results indicate that it achieves a high success rate up to 93\%.\\

\begin{table}[h!]
\centering
\caption{Model performance for Two Drone System}
\begin{tabular}{|l|c|c|}
\hline
\textbf{Training Iteration} & \textbf{Success Rate} & \textbf{Average Steps} \\
\hline
4.15 mil & 86\% & 162.89 \\
4.2 mil & 83\% & 130.08 \\
\hline
\end{tabular}
\label{tab:multi_drone_performance}
\end{table}

\begin{figure}[h!]
    \centering
    \includegraphics[width=0.8\linewidth]{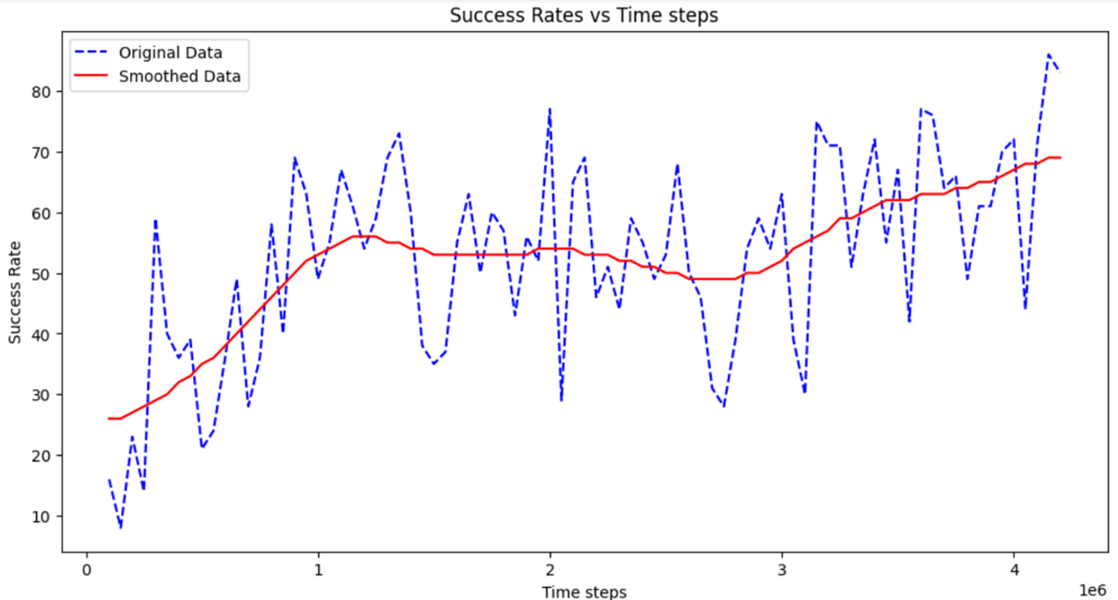}
    \caption{Evaluation graph - two drone simulation environment.}
    \label{fig:multi_drone_eval}
\end{figure}

\figurename~\ref{fig:multi_drone_eval} shows the evaluation graph for the two drone simulation environment. The model maintains a strong success rate of 86\%, with a significant decrease in the average number of steps, illustrating the benefits of decentralized search strategies.

The model excels with robust performance and high success rates in target localization. A two-drone system notably reduces steps, showcasing efficiency gains from UAV swarms. Evaluation graphs confirm adaptation across diverse environments, and training on varied datasets demonstrates strong generalization. Scaling experiments show minimal adjustments enable effective cooperation and grid map sharing. This scalability indicates potential for further reducing localization time with larger UAV swarms. \\
\section{Conclusion and Future Works}
\label{sec:CF}
This paper investigates target localization using single and multiple UAVs equipped with limited sensors in perceptually degraded environments like places without GNSS/GPS signals and poor visibility. The authors developed simulators for both single and multi-drone setups to train and evaluate their methodologies. In the single-drone model, a 93\% success rate was achieved with an average step rate of 183.82. 
When using two drones, the success rate was 86\% with an average step rate of 162.69. These results demonstrate that a coordinated UAV swarm can maintain a high success rate while reducing the steps required, showcasing the benefits of multiple drones working together in challenging environments. Future research should explore improving the obtained performances, localizing multiple emitting targets, adapting the methodology for tracking unknown mobile targets, and incorporating obstacles into the UAV-RL model state space and data readings. Additionally, implementing optimization algorithms to improve results and evaluating metrics like redundant paths and revisited cells are recommended.

\bibliographystyle{ieeetr}
\bibliography{sample}

\begin{thebibliography}{10}

\bibitem{s17102234}
J.~Aleotti, G.~Micconi, S.~Caselli, G.~Benassi, N.~Zambelli, M.~Bettelli, and
  A.~Zappettini, ``Detection of nuclear sources by uav teleoperation using a
  visuo-haptic augmented reality interface,'' {\em Sensors}, vol.~17, no.~10,
  2017.

\bibitem{Alsalam2017AutonomousUW}
B.~H.~Y. Alsalam, K.~Morton, D.~A. Campbell, and F.~Gonzalez, ``Autonomous uav
  with vision based on-board decision making for remote sensing and precision
  agriculture,'' {\em 2017 IEEE Aerospace Conference}, pp.~1--12, 2017.

\bibitem{Atif2021UAVAssistedWL}
M.~Atif, R.~Ahmad, W.~Ahmad, L.~Zhao, and J.~J. P.~C. Rodrigues, ``Uav-assisted
  wireless localization for search and rescue,'' {\em IEEE Systems Journal},
  vol.~15, pp.~3261--3272, 2021.

\bibitem{7995044}
J.~Wang, C.~Jiang, Z.~Han, Y.~Ren, R.~G. Maunder, and L.~Hanzo, ``Taking drones
  to the next level: Cooperative distributed unmanned-aerial-vehicular networks
  for small and mini drones,'' {\em IEEE Vehicular Technology Magazine},
  vol.~12, no.~3, pp.~73--82, 2017.

\bibitem{DBLP:journals/corr/abs-1802-07187}
S.~Mousavi, F.~Afghah, J.~D. Ashdown, and K.~A. Turck, ``Leader-follower based
  coalition formation in large-scale {UAV} networks, {A} quantum evolutionary
  approach,'' {\em CoRR}, vol.~abs/1802.07187, 2018.

\bibitem{article8}
H.~Peng, A.~Razi, F.~Afghah, and J.~Ashdown, ``A unified framework for joint
  mobility prediction and object profiling of drones in uav networks,'' {\em
  Journal of Communications and Networks}, vol.~20, pp.~434--442, 10 2018.

\bibitem{SHURRAB2023100867}
M.~Shurrab, R.~Mizouni, S.~Singh, and H.~Otrok, ``Reinforcement learning
  framework for uav-based target localization applications,'' {\em Internet of
  Things}, vol.~23, p.~100867, 2023.

\bibitem{article11}
Q.~Ge, C.~Wen, and S.~Duan, ``Fire localization based on range-range-range
  model for limited interior space,'' {\em Instrumentation and Measurement,
  IEEE Transactions on}, vol.~63, pp.~2223--2237, 09 2014.

\bibitem{article13}
A.~Alagha, R.~Mizouni, S.~Singh, H.~Otrok, and A.~Ouali, ``Sdrs: A stable
  data-based recruitment system in iot crowdsensing for localization tasks,''
  {\em Journal of Network and Computer Applications}, vol.~177, p.~102968, 12
  2020.

\bibitem{sutton2018reinforcement}
R.~S. Sutton and A.~G. Barto, {\em Reinforcement Learning: An Introduction}.
\newblock Adaptive Computation and Machine Learning Series, Cambridge,
  Massachusetts: The MIT Press, 2nd~ed., 2018.

\bibitem{rs16030471}
S.~Boiteau, F.~Vanegas, and F.~Gonzalez, ``Framework for autonomous uav
  navigation and target detection in global-navigation-satellite-system-denied
  and visually degraded environments,'' {\em Remote Sensing}, vol.~16, no.~3,
  2024.

\bibitem{Aasish2015NavigationOU}
C.~Aasish, E.~Ranjitha, R.~U. Razeen, R.~Bharath, and J.~L. Angelin,
  ``Navigation of uav without gps,'' {\em 2015 International Conference on
  Robotics, Automation, Control and Embedded Systems (RACE)}, pp.~1--3, 2015.

\bibitem{mnih2013playing}
V.~Mnih, K.~Kavukcuoglu, D.~Silver, A.~Graves, I.~Antonoglou, D.~Wierstra, and
  M.~Riedmiller, ``Playing atari with deep reinforcement learning,'' 2013.

\bibitem{lillicrap2019continuous}
T.~P. Lillicrap, J.~J. Hunt, A.~Pritzel, N.~Heess, T.~Erez, Y.~Tassa,
  D.~Silver, and D.~Wierstra, ``Continuous control with deep reinforcement
  learning,'' 2019.

\bibitem{schulman2017proximal}
J.~Schulman, F.~Wolski, P.~Dhariwal, A.~Radford, and O.~Klimov, ``Proximal
  policy optimization algorithms,'' 2017.

\bibitem{pleines2022generalization}
M.~Pleines, M.~Pallasch, F.~Zimmer, and M.~Preuss, ``Generalization, mayhems
  and limits in recurrent proximal policy optimization,'' 2022.

\end{thebibliography}





\end{document}